\documentclass{article} 
\usepackage[final]{colm2026_conference}

\usepackage{microtype}
\usepackage{hyperref}
\usepackage{url}
\usepackage{booktabs}

\usepackage{lineno}

\definecolor{darkblue}{rgb}{0, 0, 0.5}
\hypersetup{colorlinks=true, citecolor=darkblue, linkcolor=darkblue, urlcolor=darkblue}

\usepackage{amsfonts}       
\usepackage{nicefrac}       
\usepackage{microtype}      
\usepackage{xcolor}         
\usepackage{graphicx}
\usepackage{wrapfig}

\usepackage{amsmath}
\usepackage{amssymb}
\usepackage{mathtools}
\usepackage{amsthm}

\usepackage{listings}
\definecolor{codegreen}{rgb}{0,0.6,0}
\definecolor{codegray}{rgb}{0.5,0.5,0.5}
\definecolor{codepurple}{rgb}{0.58,0,0.82}
\definecolor{backcolour}{rgb}{0.95,0.95,0.92}
\theoremstyle{plain}
\newtheorem{theorem}{Theorem}[section]

\theoremstyle{definition}
\newtheorem{definition}[theorem]{Definition}

\theoremstyle{remark}

\definecolor{myorange}{HTML}{ee8e2c}
\definecolor{mygreen}{HTML}{379d4e}
\definecolor{myred}{HTML}{f21c19}

\title{MAPLE: Metadata Augmented Private Language Evolution}

\author{Eli Chien\textsuperscript{*} \\
  National Taiwan University \\
  \texttt{elichien@ntu.edu.tw} \\
  \And
  Yuzheng Hu\textsuperscript{* \dag} \\
  University of Illinois Urbana-Champaign \\
  \texttt{yh46@illinois.edu} \\
  \And
  Ryan McKenna \\
  Google Research \\
  \texttt{mckennar@google.com} \\
  \And
  Shanshan Wu \\
  Google Research \\
  \texttt{shanshanw@google.com} \\
  \And
  Zheng Xu\textsuperscript{\ddag} \\
  Meta \\
  \texttt{zhx@meta.com} \\
  \And
  Peter Kairouz \\
  Google Research \\
  \texttt{kairouz@google.com} \\
}

\begin{document}

\ifcolmsubmission
\linenumbers
\fi

\maketitle
\begingroup
\renewcommand{\thefootnote}{\fnsymbol{footnote}}
\footnotetext[1]{Equal contribution.}
\footnotetext[2]{Work done while interning at Google Research.}
\footnotetext[3]{Work done at Google Research.}
\endgroup

\begin{abstract}

Differentially private (DP) fine-tuning of large language models (LLMs) requires massive compute and full model access, which rules out state-of-the-art proprietary APIs for general users. Generating DP synthetic data offers a practical workaround. This approach also allows for transparent exploratory data analysis and arbitrary reuse across downstream tasks, sidestepping the rigid constraints of a model's parameter space. Private Evolution (PE) provides a promising API-based framework for generating this data, but its success relies heavily on initialization. If the private data distribution falls too far outside the foundation model's pre-training priors -- a common issue in highly specialized domain -- PE struggles to align with the target data. This misalignment causes poor convergence, degraded utility, and wasted API calls. To solve this initialization bottleneck, we introduce Metadata Augmented Private Language Evolution (MAPLE). MAPLE extracts DP tabular metadata and uses in-context learning to firmly ground the initial synthetic distribution in the target domain. Our evaluations on domain-specific text generation tasks show that MAPLE yields a strictly better privacy-utility trade-off, converges significantly faster, and sharply reduces API costs compared to baseline PE methods. Our code is available at {\url{https://github.com/elichien-google/MAPLE}}.

\end{abstract}

\section{Introduction}\label{sec:intro}

The remarkable capabilities of modern artificial intelligence, particularly Large Language Models (LLMs), are driven by training on vast quantities of user-generated text~\citep{ouyang2022training,touvron2023llama,liang2023holistic}. Such data, ranging from mobile keyboard inputs~\citep{xu2023federated,zhang2025synthesizing} to sensitive medical records~\citep{rumshisky2016predicting} and recommendation histories~\citep{karatzoglou2017deep,zhang2019deep}, form the backbone of model performance. However, this reliance on large-scale, user-generated data necessitates rigorous privacy safeguards; LLMs are known to memorize their training data, posing significant risks of sensitive information leakage~\citep{carlini2021extracting,lukas2023analyzing,wang2023decodingtrust}. 

Differentially Private (DP) synthetic text has emerged as a compelling approach to mitigating these risks. Due to the post-processing property of DP~\citep{dwork2014algorithmic}, synthetic datasets can be reused across diverse downstream tasks without incurring additional privacy costs if the training data for generating DP synthetic data are not used again in downstream tasks. Furthermore, sharing synthetic data enables practitioners to perform transparent exploratory data analysis without the opaque constraints of a model's parameter space. Consequently, generating high-utility DP synthetic data has become a central goal, often described as the ``holy grail'' of privacy-preserving machine learning~\citep{hu2024sok,lin2024differentially}.

\begin{figure*}[t]
    \centering 
    \includegraphics[width=0.99\textwidth]{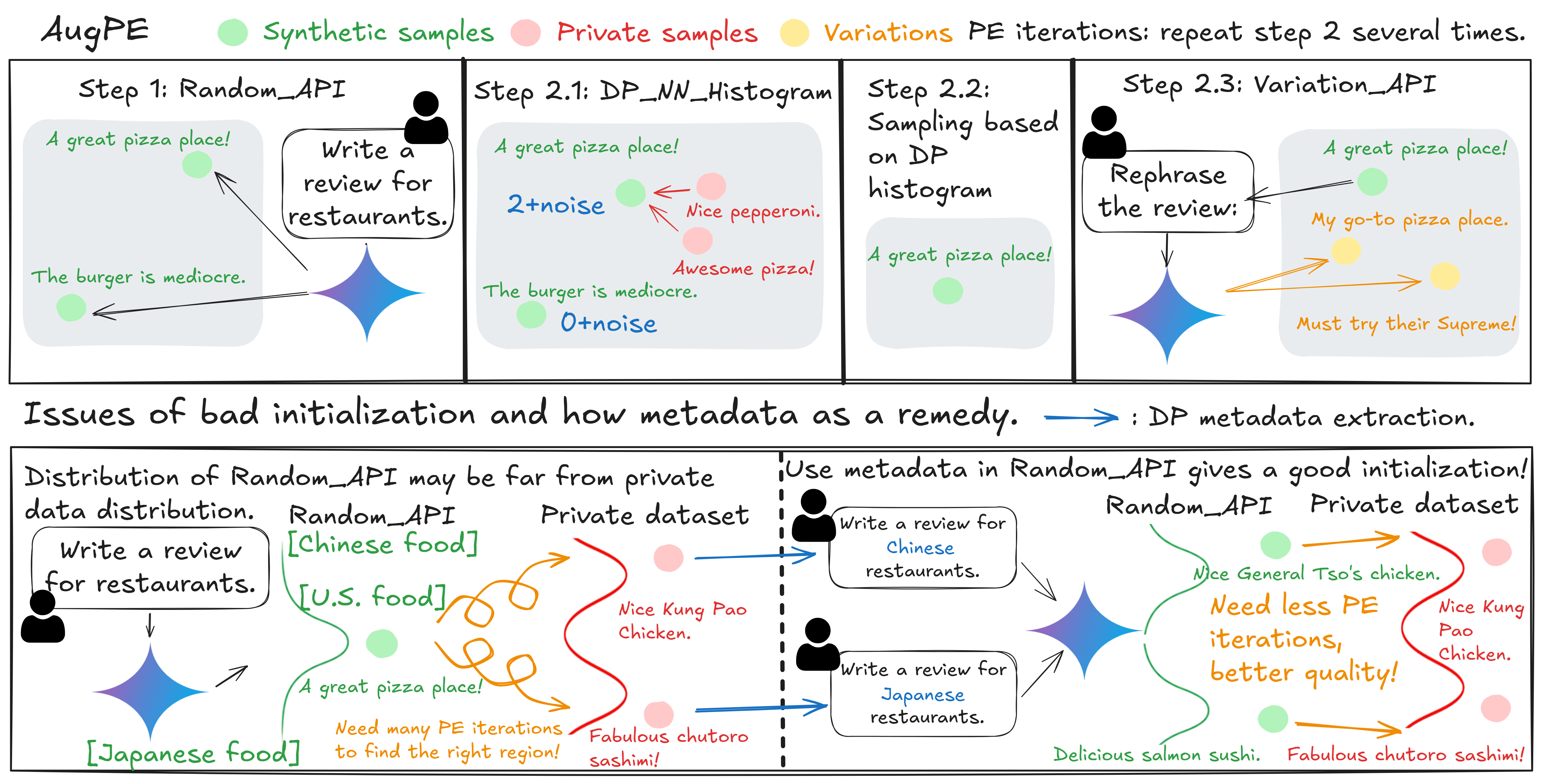}
    \caption{Illustration of AugPE and its limitations. \textbf{Top}: Overview of AugPE, adapted from~\citet{xie2024differentially}. AugPE first uses \texttt{RANDOM\_API} to generate \textcolor{mygreen}{synthetic samples} with a \textit{data-independent} prompt (Step 1). It then iteratively refines these \textcolor{mygreen}{synthetic samples} toward the \textcolor{myred}{private samples}. At each iteration, AugPE selects \textcolor{mygreen}{synthetic samples} that receive high nearest-neighbor votes from \textcolor{myred}{private samples} using a differentially private voting histogram in an embedding space (Step 2.1 and 2.2). The selected \textcolor{mygreen}{synthetic samples} are subsequently paraphrased via \texttt{VARIATION\_API} to produce \textcolor{myorange}{new synthetic samples} (Step 2.3). This process is repeated for multiple rounds. \textbf{Bottom}: When the initial distribution induced by \texttt{RANDOM\_API} is poorly aligned with the \textcolor{myred}{private samples}, AugPE may require many iterations to reach the region of the private data distribution. To address this limitation, we propose incorporating metadata into the \texttt{RANDOM\_API} prompt, making the initialization \textit{data-dependent} and better aligned for the PE process. Details on differentially private metadata extraction are in Appendix~\ref{apx:step0} and Figure~\ref{fig:maple}.}
    \label{fig:prelim}
    \vspace{-0.3cm}
\end{figure*}

The most direct approach to generating DP synthetic text is to DP-finetune a generative language model on private corpora~\citep{yue2023synthetic,kurakin2023harnessing,yu2024privacy}. However, several practical constraints often render this approach prohibitive. First, many state-of-the-art LLMs, such as Gemini 3~\citep{google2025gemini3}, GPT 5.2~\citep{openai2025gpt52}, 
~\citep{anthropic2026claude46}, are accessible only through proprietary APIs. While standard finetuning APIs are available for a limited set of models \citep{openai2023finetuning}, DP-finetuning requires specialized gradient-level updates (per-sample clipping and noise injection) that are not supported by any commercial API provider to date.

Second, even when high-performance open-source models like Llama~\citep{grattafiori2024llama} or Qwen~\citep{qwen2.5} are available, DP-finetuning remains computationally expensive and challenging to implement~\citep{sinha2025vaultgemma}. This burden is particularly acute for data holders in resource-constrained settings, including medical clinics, financial institutions, and individual end-users. 
Moreover, similar difficulties arise even in large-scale industrial environments when private data must be processed within Trusted Execution Environments (TEEs)~\citep{sabt2015trusted} or under the constraints of federated learning~\citep{kairouz2021advances}, where access to system resources and gradient-level operations is inherently restricted.

Recently, the Private Evolution (PE) framework~\citep{lin2024differentially} has emerged as a promising alternative in settings where DP-finetuning is prohibitive. PE requires only API access to foundation models while achieving a privacy-utility trade-off competitive with state-of-the-art methods~\citep{xie2024differentially}. However, the effectiveness of PE critically depends on the quality of its initialization, which is obtained via direct prompting (Figure~\ref{fig:prelim}). In particular, PE implicitly assumes that the initial unconditional generations are reasonably aligned with the target private data distribution. This assumption can break down when the private data distribution deviates substantially from the LLM’s pre-training prior. These observations naturally raise a central research question: ``\textit{How can we design effective initializations for PE that improve the quality of DP synthetic data, especially under distributional mismatch?}''

We address this challenge by proposing \textbf{Metadata Augmented Private Language Evolution (MAPLE)}, a framework that leverages structural metadata to guide the evolution process while maintaining end-to-end differential privacy guarantees. A key feature of MAPLE is that it does not rely on pre-defined metadata being available in the raw private data; instead, we design a differentially private pipeline that extracts and synthesizes tabular metadata directly from the raw text. To further improve synthetic data quality, MAPLE exploits the in-context learning capabilities of LLMs by incorporating a small number of in-context examples into the prompt. 
Extensive experiments on two domain-specific datasets demonstrate that MAPLE consistently outperforms AugPE~\citep{xie2024differentially}, the current state-of-the-art PE-based method, across all privacy regimes and evaluation metrics. Notably, MAPLE converges using substantially fewer API calls than AugPE. This reduction in query complexity leads to a more cost-effective and economically efficient solution for generating high-quality DP synthetic data, without requiring model finetuning.

\begin{figure*}[t]
    \centering 
    \includegraphics[width=0.99\textwidth]{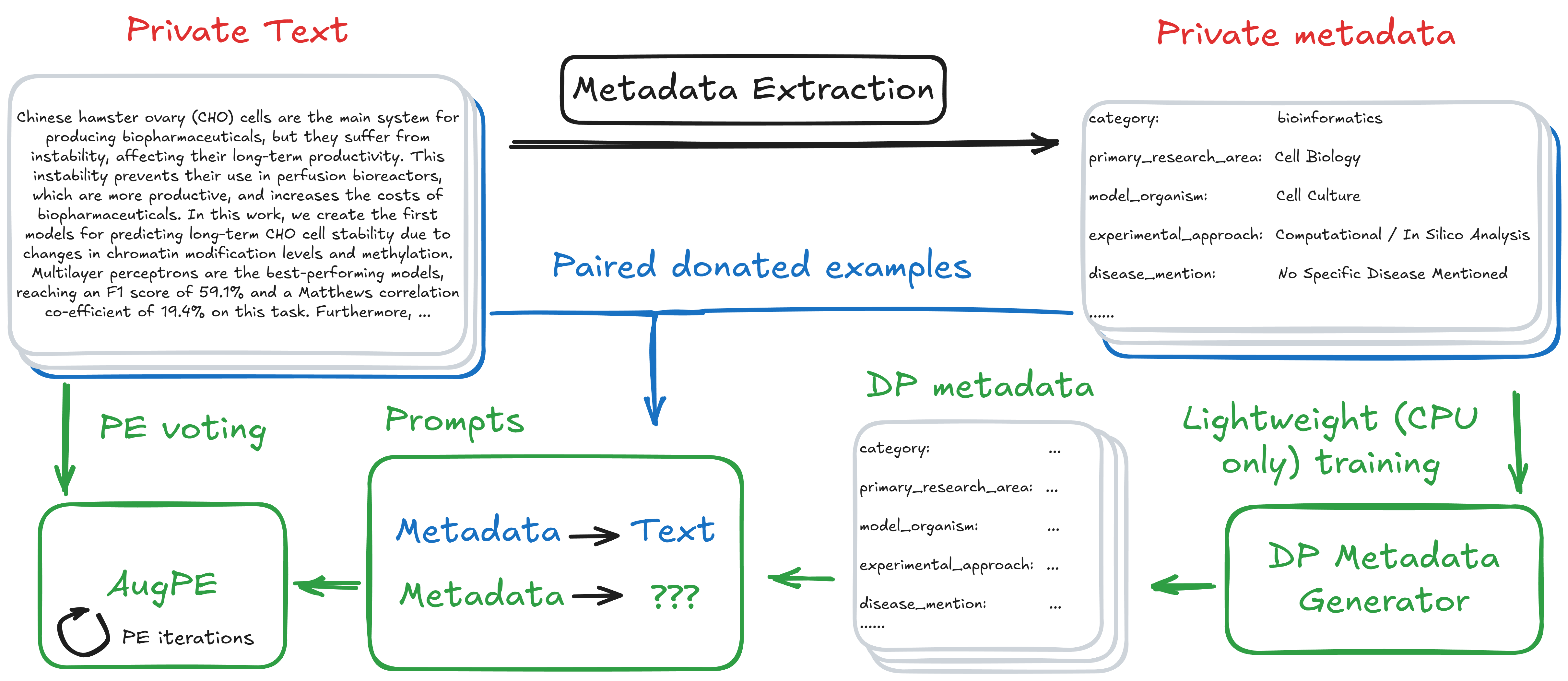}
    \caption{Overview of MAPLE. We first extract metadata in tabular format, either based on a designed schema or adapt it from the dataset whenever it is given. Then we train a light-weighted (with CPU only) DP metadata generator via the state-of-the-art approach, such as AIM~\citep{mckenna2022aim}. Next, we compose the prompt for \texttt{RANDOM\_API} with both DP metadata and a few donated (metadata, text) pairs as in-context examples. As we demonstrate in the ablation study in Section~\ref{sec:exp}, our in-context example design is crucial for fully leveraging the in-context learning capability of LLMs. After the initialization, we refine it by the PE (i.e., AugPE) for the final DP synthetic dataset.}
    \label{fig:maple}
\end{figure*}

\section{Preliminaries}\label{sec:prelim}

\textbf{Differential Privacy (DP).}  DP~\citep{dwork2006calibrating} is the gold standard for privacy, which is not only adopted by the U.S. government~\citep{abowd2018us} but also by the industry~\citep{ding2017collecting,thakurta2017learning,xu-etal-2023-federated}. It provides a rigorous guarantee that limits what an adversary can infer about any single user's data from an algorithm's output. 

\begin{definition}\label{def:DP}
    A randomized algorithm $\mathcal{M}$ is $(\varepsilon,\delta)$-DP if for any two adjacent datasets $\mathcal{D},\mathcal{D}^\prime$ (i.e., adding or removing one sample) and any subset of possible output $S$, we have
    \begin{align}
        & \mathbb{P}\left[ \mathcal{M}(\mathcal{D}) \in S\right] \leq e^\varepsilon \mathbb{P}\left[ \mathcal{M}(\mathcal{D}^\prime) \in S\right] + \delta.
    \end{align}
\end{definition}
$\varepsilon\in [0,\infty)$ is often referred to as the privacy loss, where a smaller $\varepsilon$ implies a stronger privacy guarantee. $\delta$ represents a small failure probability, which is often set to be the reciprocal of the dataset size.

DP has many desirable properties for AI applications, especially private synthetic data generation. The \textit{post-processing} property ensures that any post-processing of an $(\varepsilon,\delta)$-DP mechanism does not incur additional privacy loss~\citep{dwork2014algorithmic} when the private dataset is not further touched. Moreover, the (sequential) composition theorem~\citep{kairouz2015composition,mironov2017renyi} characterizes the overall privacy loss of the composition of multiple DP mechanisms.

\textbf{Private Evolution (PE).} PE~\citep{lin2024differentially}, and its text-specific instantiation AugPE~\citep{xie2024differentially}, are iterative algorithms that utilize two types of API calls to LLMs. The first is \texttt{RANDOM\_API}, which generates an initial set of synthetic text with a data-independent prompt (Step 1)\footnote{While~\citep{xie2024differentially} utilizes class labels available in the OpenReview dataset, they treat this information as public and do not ensure corresponding DP protection.}. The second is \texttt{VARIATION\_API}, which paraphrases a given text to produce a similar synthetic sample (Step 2.3). An overview of AugPE is shown in Figure~\ref{fig:prelim}. Notably, a key assumption underlying the effectiveness of PE-based methods is that the initial synthetic samples produced by \texttt{RANDOM\_API} are reasonably aligned with the private data~\citep{lin2024differentially}. 
Indeed, when most private data lie in the low probability region of the initial synthetic distribution, the initial samples can be far from the private data, causing PE to struggle to make progress.
We illustrate this effect in the bottom row of Figure~\ref{fig:prelim}: repeatedly paraphrasing a review about ``pizza'' is unlikely to produce content related to Chinese or Japanese restaurants via \texttt{VARIATION\_API}. 
As a result, a poorly aligned initialization can require many more PE iterations to achieve satisfactory synthetic data quality. 
This not only increases cost due to additional API calls, but also leads to a worse privacy-utility trade-off, since each iteration incurs additional privacy loss from calling \texttt{DP\_NN\_Histogram}.

\section{MAPLE: Metadata Augmented Private Language Evolution}\label{sec:MAPLE}
To address the poor initialization of PE, we propose to provide LLMs with additional information about the private dataset by incorporating metadata into the \texttt{RANDOM\_API} prompt; see Figure~\ref{fig:maple} for an overview. The algorithm is provided in Appendix~\ref{apx:maple_procedure}.

\textbf{Step 0: Pre-processing.} We first extract metadata from the raw private text corpus. This step is crucial as it not only prevents us from assuming the existence of the auxiliary information in the private dataset, but also offers an opportunity to leverage domain knowledge for what metadata should be extracted. Following~\citet{hu2025actg}, we choose to extract the metadata with a structured tabular schema. The schema is rich and multi-attribute with fixed options per attribute. The resulting metadata is in a tabular format and automatically annotated by LLMs. We refer interested readers to Appendix~\ref{apx:step0} for further details on this pre-processing step.

\textbf{Step 1: DP tabular metadata.} While providing metadata in the prompt of \texttt{RANDOM\_API} lets LLMs know more information about the private dataset, it also breaches the privacy of the private dataset. We need to ensure a proper DP guarantee for our metadata, which is also the reason why we restricted the metadata to the tabular format. Generating DP synthetic tabular data is a well-studied problem. In particular, the lightweight algorithm AIM~\citep{mckenna2022aim} is shown to offer strong tabular generation quality and only requires CPU training. This is particularly suitable for our resource-constrained scenario. We leverage AIM as our DP metadata generator, trained with private metadata.

\textbf{Step 2: Prompt for \texttt{RANDOM\_API}.} To fully utilize the generation quality of powerful LLMs, we further utilize their in-context learning capability via few-shot examples. Before providing the synthetic metadata for text generation, we give a few (metadata, text) pairs as in-context examples. In practice, these non-private in-context examples can be either obtained from donated data from users who opt out of the privacy requirement or curated by the data holder~\citep{nasr2023effectively,wang2024role}. They are therefore outside the protected private dataset by construction. In our ablation study in Section~\ref{sec:exp}, we found that providing such (metadata, text) pairs as in-context examples is crucial. Utilizing merely metadata or example text only provides a marginal improvement over the vanilla AugPE baseline.

\textbf{Step 3: AugPE. }After obtaining the initial DP synthetic text dataset, we further refine it with a few AugPE iterations. As we will see in the experiment, starting with our enhanced initial DP synthetic text dataset not only results in better final synthetic data quality, but also converges with many fewer iterations. Interestingly, we find that it is still crucial to utilize a few AugPE iterations for the best overall results.

\textbf{Privacy guarantee. }MAPLE protects the private training corpus through the composition of two DP mechanisms: AIM for synthetic metadata generation and the Gaussian mechanism applied to the PE NN histograms. The LLM calls use only DP synthetic metadata, non-private in-context examples, or intermediate synthetic text, and are therefore post-processing of the DP mechanisms. Hence, the final synthetic text dataset satisfies an end-to-end DP guarantee with respect to all private training records.

\section{Related Works}\label{sec:relate}

\textbf{DP synthetic text. }Methods for DP synthetic text generation are primarily divided into two paradigms: DP-finetuning and Private Evolution (PE). DP-finetuning approaches optimize pre-trained LLMs on private corpora using differentially private optimizers, most notably DP-SGD~\citep{abadi2016deep, yue2023synthetic, carranza2024synthetic, ochs2025private}. Despite their effectiveness, these methods are often prohibitive when using closed-source models like GPT-4 or when operating within resource-constrained environments such as Trusted Execution Environments (TEEs)~\citep{sabt2015trusted}.

In contrast, the PE paradigm~\citep{lin2024differentially} relies exclusively on LLM inference, making it better suited for black-box or compute-limited scenarios. Recent advancements in PE have focused on algorithmic refinements~\citep{zou2025contrastive} and extensions to federated learning frameworks~\citep{hou2024pretext, hou2025private}. These developments are orthogonal to our work; while prior research optimizes the iterative evolution process, MAPLE focuses on enhancing the initialization phase through DP-protected metadata and in-context learning.

Another emerging category is private inference, which injects DP noise directly into the next-token logits during the prompting process~\citep{amin2024private}. However, private inference typically incurs higher privacy costs as the length of the generated sequence increases, and will feed private text to LLMs for obtaining logits. We view PE and private inference as complementary approaches with distinct use cases. Finally, our work bridges the gap between text and structured data by incorporating principles from DP tabular synthesis~\citep{hu2024sok} into the text domain. For a more comprehensive overview of the broader DP synthetic data landscape, we refer readers to recent surveys by~\citet{hu2024sok} and~\citet{ponomareva2025dp}.

\textbf{Leveraging metadata. }Recently, several works have been proposed to improve DP synthetic text generation via the use of metadata. ~\citet{tansynthesizing,yu2024privacy} improves the DP finetuning-based approach by distribution alignment defined on metadata in an \textit{a priori} and \textit{post-hoc}, respectively. ~\citet{hu2025actg} further extends the idea of~\cite{tansynthesizing} by introducing RL finetuning to have a better controllability for conditional generation. Very recently,~\citet{sun2025dpga} also proposed to leverage the idea of private metadata as part of their improved design for PE. However, they do not explore the in-context learning capability of LLMs as we do, but rather provide it directly in the prompts of API calls. As we have observed in our ablation study, providing both metadata and in-context examples is crucial for the best results.

\section{Experiments}\label{sec:exp}

\begin{figure*}[t]
    \centering 
    \includegraphics[width=0.99\textwidth]{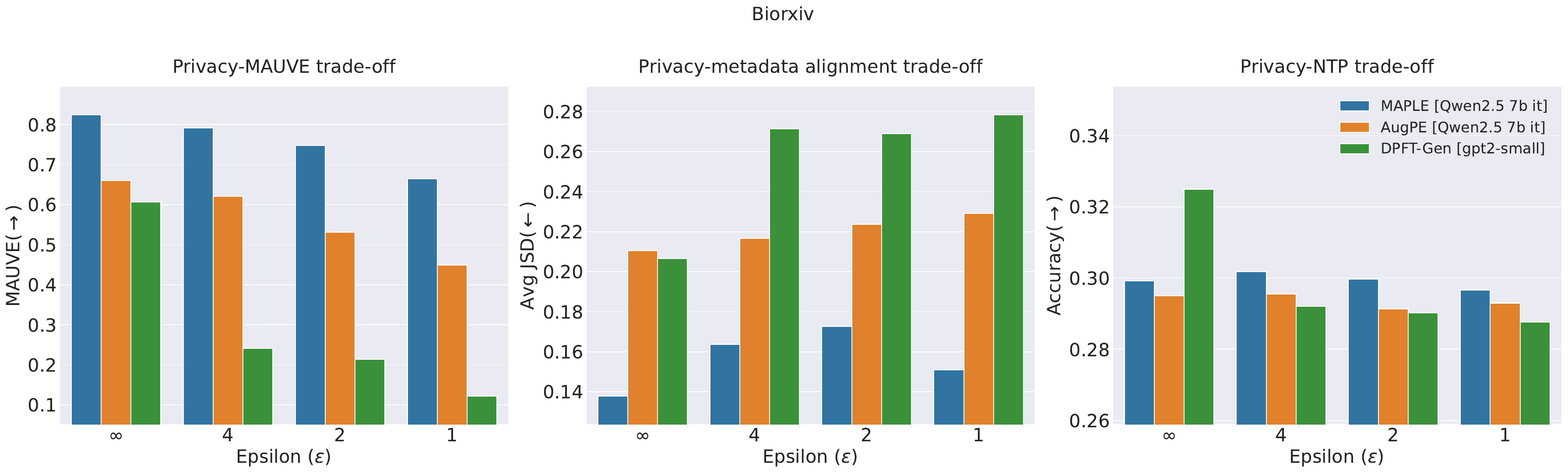}
    \includegraphics[width=0.99\textwidth]{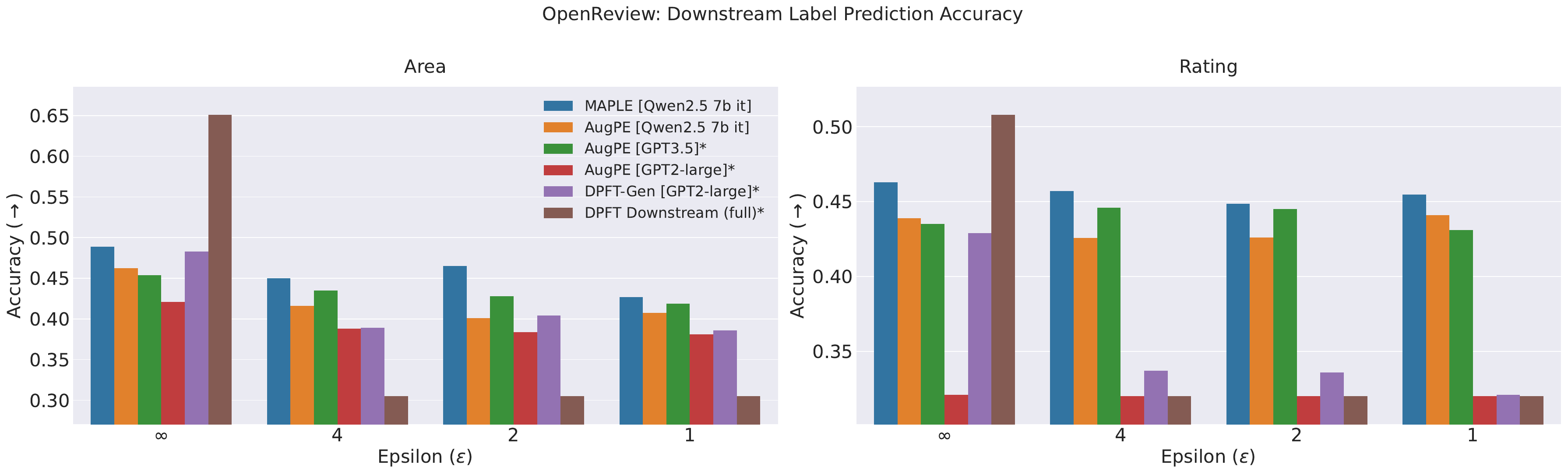}
    \caption{Main results. Top: Biorxiv datasets. The utility metrics are MAUVE score, average JSD on all metadata annotated by a powerful LLM, and NTP accuracy for training a bert-small model using synthetic data. Bottom: OpenReview datasets. The utility metrics are downstream prediction accuracy for using synthetic data to train a roberta-base model following the same setting of~\citet{xie2024differentially}, where the predicted labels are the area and rating of the review. The mark $*$ indicates the results are directly taken from~\citet{xie2024differentially}.}
    \label{fig:mainexp}
\end{figure*}

\textbf{Datasets.} We conduct experiments on two challenging domain-specific datasets. \textit{bioRxiv}~\citep{hou2025private} consists of a corpus of 29k scientific abstracts from preprints, with an average length of about 300 tokens. \textit{OpenReview}~\citep{xie2024differentially} contains roughly 8.4k peer reviews for ICLR 2023 submissions collected from the
OpenReview website~\footnote{\url{https://openreview.net/group?id=ICLR.cc/2023/Conference}}. 
Compared to commonly used benchmarks such as Yelp~\cite{yelp_dataset} in prior works~\citep{yue2023synthetic,xie2024differentially,tansynthesizing}, these datasets pose a more challenging synthetic text generation task. They are not only domain-specific, but also have longer average lengths (for example, Yelp has an average of 173 tokens). We provide further details for all datasets in Appendix~\ref{apx:expdetail}, and include an additional experiment on IMDB in Appendix~\ref{apx:imdb_init}.

\textbf{Evaluation metrics. }We evaluate the synthetic data quality from multiple facets, capturing various aspects of the synthetic text. For \textit{bioRxiv}, we assess the general fidelity by MAUVE~\citep{pillutla2021mauve} as in prior works~\citep{xie2024differentially,hu2025actg}. It quantifies the semantic similarity between the synthetic and private text at the distribution level. We chose sentence-t5-base~\citep{ni2022sentence} as the embedding model for MAUVE calculation, which better captures the text semantics compared to weaker embedding models such as stsb-roberta-base-v2~\citep{reimers-2019-sentence-bert} or all-MiniLM-L6-v2\footnote{\url{https://huggingface.co/sentence-transformers/all-MiniLM-L6-v2}}. We also examine how the distribution of the annotated metadata from synthetic text aligns with the private one via the Jensen-Shannon distance between them. It provides a more fine-grained assessment of the particular aspects of the synthetic text compared to the general semantic meaning provided by MAUVE. For utility, we report the next token prediction (NTP) accuracy on a downstream generation task, where we trained a downstream GPT-2 small model~\citep{radford2019language} on the synthetic text and evaluated the NTP accuracy on the test set. For \textit{OpenReview}, we follow the same setting as in~\citet{xie2024differentially} to evaluate the downstream classification accuracy, where the labels are review recommendations (5 classes) and areas (12 classes). We finetune the RoBERTa-base~\citep{liu2019roberta} as text classifiers. Notably, prior work~\citep{xie2024differentially} treats these downstream labels as \textit{public} and uses them from conditional generation in AugPE directly. In contrast, our MAPLE framework treats these labels as \textit{private} metadata, which better aligns with the common practice.

\textbf{Baselines. }We mainly compare with two baselines: AugPE~\citep{xie2024differentially} and DP finetuned generator (DPFT-Gen)~\citep{yue2023synthetic}. The setup of these baselines follows the implementation of~\citet{xie2024differentially}. For OpenReview, we directly take the results from~\citet{xie2024differentially}. For the other datasets that are not tested in~\citet{xie2024differentially}, we mainly follow their hyperparameter setting and optimized hyperparameters whenever necessary. Note that while there are some recent works that improve AugPE iterations~\citep{zou2025contrastive,sun2025dpga}, our improvement on initialization is orthogonal to them and may be combined for even better results. Since we focus solely on demonstrating the effect of better initialization for AugPE via metadata, we choose the vanilla AugPE as our primary baseline.

\textbf{Implementation details. }We choose \texttt{Qwen2.5-7B-Instruct}\footnote{\url{https://huggingface.co/Qwen/Qwen2.5-7B-Instruct}}~\citep{qwen2.5} as the base LLMs in our implementation of AugPE and MAPLE. Notably, the official AugPE implementation~\footnote{\url{https://github.com/microsoft/DPSDA}} uses the HuggingFace library as its backbone, which is sub-optimal in efficiency compared to the recent efficient LLM serving library such as vLLM~\citep{kwon2023efficient}. This effect becomes even more critical in our MAPLE framework, as the in-context examples we provide in the prompt can have more than 3000 tokens. As a result, we re-implement AugPE with vLLM and use it as the building block of our MAPLE framework. This allows us to have a rigorous ablation study on the effect of initialization compared to taking the official implementation directly.

For all datasets, we fix the size of the donated text set to be $50$, which is sampled from the validation set. Note that compared to both the targeted size of the synthetic dataset (2k or 5k), and the private training set size, assuming $50$ donated examples is a practical setting (less than $3\%$ at most). For each synthetic metadata, we choose the top $10$ donated texts with the smallest Hamming distance on the metadata. We found that providing more than $10$ in-context examples results in diminished return. For the metadata extraction, we utilize \texttt{gemini-2.5-flash-lite}\footnote{\url{https://ai.google.dev/gemini-api/docs/models\#gemini-2.5-flash-lite}} as the metadata annotator. Further discussion is deferred to Appendix~\ref{apx:mapledetail}.

\textbf{Privacy accounting. }We evaluate all methods under privacy budgets $\varepsilon \in \{1, 2, 4, \infty\}$ with $\delta = 1/n$, where $n$ is the private training set size. The total privacy cost of MAPLE comprises the DP metadata generator (AIM) and the PE voting iterations. We allocate the budget via zero-concentrated DP (zCDP) accounting~\citep{bun2016concentrated}: the total $(\varepsilon, \delta)$ budget is converted to a zCDP parameter $\rho$, split as $\rho_{\text{AIM}} = 0.1\rho$ and $\rho_{\text{PE}} = 0.9\rho$. Each of the $T$ voting iterations adds independent Gaussian noise $\mathcal{N}(0, \sigma^2)$ to the nearest-neighbor voting histogram. By zCDP composition, the per-iteration cost is $\rho_{\text{PE}}/T$, yielding $\sigma = \sqrt{T/(2\rho_{\text{PE}})}$. For AugPE, the entire budget is allocated to PE voting.

\textbf{Hyperparameters and hardware. }The main hyperparameter is the number of private evolution iterations $T$. For both MAPLE and AugPE, we first run the non-private version with $T=10$ to identify the iteration at which MAUVE peaks (Figure~\ref{fig:biorxiv_peiter}), then search locally around that value in the private setting. 
Unlike the non-private case where a single run with large $T$ suffices, each choice of $T$ under DP requires a separate run with a different noise multiplier $\sigma$, making exhaustive search over $T$ computationally expensive.
Following common practice in DP synthetic data generation~\citep{xie2024differentially}, we report single-run results.
All experiments were conducted on NVIDIA A100 40GB GPUs.

\subsection{Main results}\label{sec:exp_main}

\begin{figure*}[t]
    \centering 
    \includegraphics[width=0.99\textwidth]{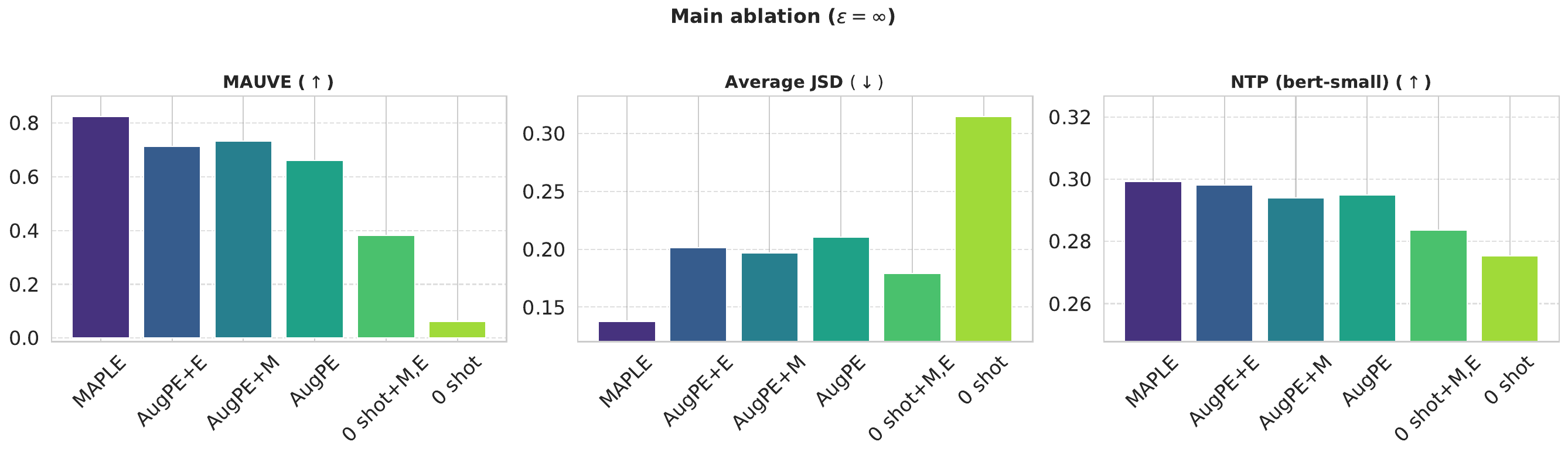}
    \caption{Ablation study on Biorxiv dataset. +M: leveraging only metadata in the prompt. +E: leveraging only in-context examples in the prompt. MAPLE: leverage both in the prompt. 0 shot: direct prompting without PE iterations.}
    \label{fig:mainablation}
\end{figure*}

We first compare the end-to-end performance of MAPLE with the baselines on different datasets. MAPLE consistently outperforms AugPE across all datasets, metrics, and privacy constraints. It validates the effectiveness of leveraging metadata and in-context learning capability in MAPLE. We also observe that MAPLE can outperform the DP finetuning generator approach. This observation is aligned with~\citet{xie2024differentially}, where the PE-based approach may offer a better privacy-utility trade-off than the DP finetuned generator. Surprisingly, we observe that MAPLE with Qwen 2.5 7b can even surpass the performance of AugPE with GPT3.5 on the Openreview dataset, despite GPT3.5 being a much larger LLM. This result further highlights the importance of our design in MAPLE. 

\subsection{Ablation studies}\label{sec:exp_ablation}
\textbf{MAPLE core components.} We conduct a series of controlled ablation studies to isolate the contributions of different components in MAPLE. Specifically, we consider the following variants. (1) \textbf{AugPE+M}: only synthetic metadata is included in the \texttt{RANDOM\_API} prompt. (2) \textbf{AugPE+E}: only 10 uniformly sampled donated examples are included in the prompt (note metadata is unavailable for similarity-based selection in this setting.) Recall that MAPLE leverages both metadata (M) and donated text (E), which can be viewed as \textbf{AugPE+M,E}. 
We additionally evaluate the quality of the initialization without PE iterations. (3) \textbf{0 shot}: the initial synthetic dataset in AugPE, without PE voting or iterative refinement. (4) \textbf{0 shot+M,E}: the MAPLE initialization, using both metadata and in-context examples, without PE voting or iterative refinement. 
The latter two ablation studies assess the importance of PE iterations and whether a good initialization alone is sufficient to produce high-quality synthetic data. 

Figure~\ref{fig:mainablation} presents the comparison under $\varepsilon=\infty$; we observe that using either metadata or in-context examples alone yields limited improvement over AugPE. It is crucial to leverage both metadata and in-context examples to fully exploit LLM capabilities. We also confirm the necessity of PE iterations. While incorporating in-context examples in the 0-shot setting substantially improves generation quality, it still falls short of the performance achieved by MAPLE and even AugPE. These results show that PE iterations, metadata, and in-context examples are all essential for achieving high-quality synthetic data. The poor 0-shot results also suggest that general pretraining is insufficient to match the specific long-form distributions of bioRxiv abstracts without dataset-specific initialization.

This conclusion holds under finite privacy budgets: we compare MAPLE with AugPE+E on bioRxiv under $\varepsilon=1$ and $\varepsilon=4$. MAPLE achieves MAUVE scores of $0.6846$ and $0.7388$, respectively, compared with $0.4915$ and $0.5732$ for AugPE+E. 
This confirms that metadata contributes substantial gains beyond donated examples alone under varying DP constraints.

\begin{figure}[t]
  \begin{center}
    \includegraphics[width=0.7\textwidth]{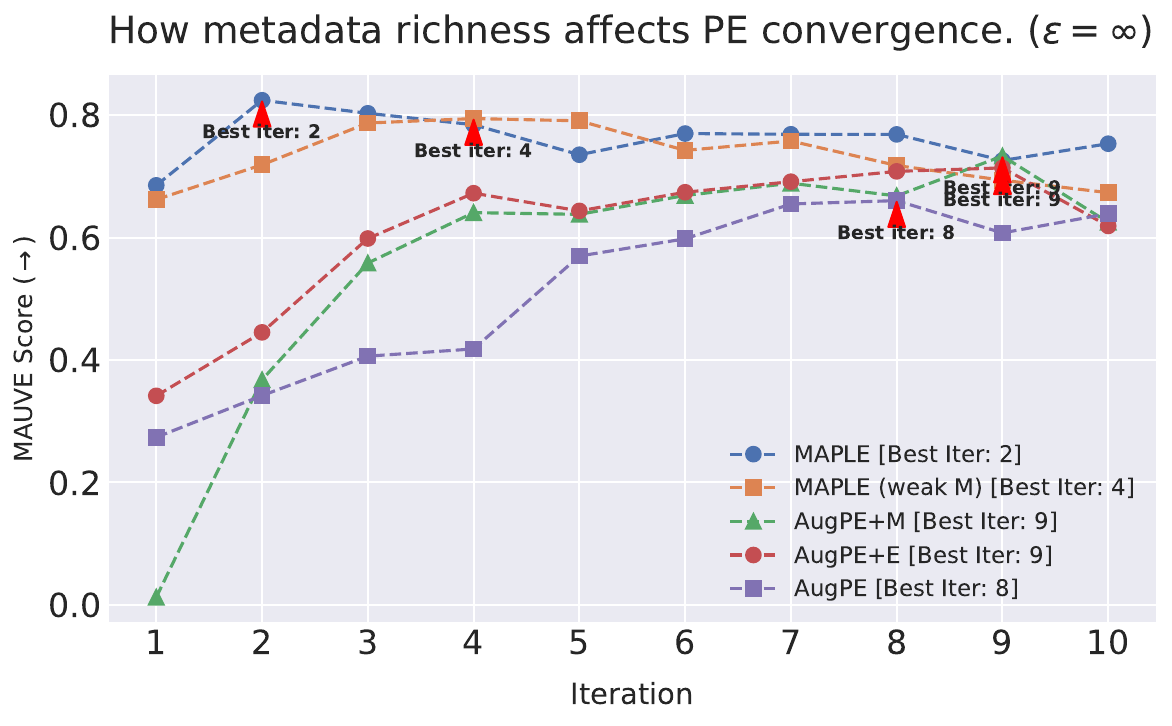}
  \end{center}
  \caption{How metadata richness affects PE convergence, measured by MAUVE score on the Biorxiv dataset. Weak M uses only two metadata attributes from the full set.}
  \label{fig:biorxiv_peiter}
\end{figure}

\textbf{Metadata richness.} We examine how the amount of information provided in the API prompt affects PE convergence, which can be viewed as an ablation on metadata richness (Figure~\ref{fig:biorxiv_peiter}). AugPE requires 8 iterations to reach its best generation quality. In contrast, MAPLE converges in only 2 iterations while achieving substantially higher quality. This not only reduces the cost of API calls, but also implies that less DP noise is needed for each PE iteration due to the DP composition theorem. Interestingly, we find that using weaker metadata with 2 attributes requires 4 iterations, more than the full metadata (9 attributes), but still far fewer than AugPE. 
These results indicate that richer metadata leads to faster PE convergence and improved peak performance. Under the DP budgets in Figure~\ref{fig:mainexp}, MAPLE uses $T=3,3,2$ PE iterations for $\varepsilon=4,2,1$, while AugPE uses $T=9,9,8$. Since other PE parameters are held fixed, this gives roughly a \textbf{$3$--$4\times$ reduction} in API calls.

\section{Conclusion and Future Directions}

In this work, we introduced MAPLE, a framework designed to overcome the initialization bottleneck in the Private Evolution (PE) algorithm. By synergistically leveraging DP-extracted tabular metadata and in-context examples, MAPLE provides a more informed starting point for synthetic data generation. Our extensive experiments and ablation studies underscore the critical role of initialization in API-based DP frameworks and demonstrate that MAPLE significantly improves convergence and utility while maintaining strict privacy guarantees.

Looking ahead, several promising avenues for future research remain:
\begin{itemize}
    \item \textbf{Extending Beyond Initialization:} In its current form, MAPLE primarily focuses on optimizing the initial distribution (i.e., \texttt{RANDOM\_API}). However, the insights derived from metadata and in-context learning could inherently benefit other modules within the PE pipeline, such as the \texttt{VARIATION\_API}. While we maintained a controlled experimental setup to isolate the effects of initialization, investigating how these components can assist in the iterative evolution process could further enhance the privacy-utility trade-off.
    \item \textbf{Selection Strategy:} Existing PE implementations, starting from the original PE work, typically select the top-$K$ samples from the DP voting histogram at each iteration. This simple strategy may reduce diversity when many similar samples receive high votes. Designing and testing diversity-aware selection strategies is an interesting direction for future work.
    \item \textbf{Cross-Modal Generalization:} A natural extension is to adapt the MAPLE framework to other data modalities, including images, tabular data, and multimedia (audio/video). Determining how to extract representative and helpful metadata under DP constraints for non-textual data remains a compelling challenge for the PE-based approach.
    \item \textbf{Fully Private Exemplar Selection:} Although our current implementation utilizes a small set of ``donated'' data points for in-context learning, this requirement can be further relaxed. When public or donated examples are unavailable, private-prediction-based algorithms such as~\citet{amin2024private} could be used to ``rewrite'' private examples with a formal DP guarantee, at the cost of additional privacy budget. We expect minor shifts between donated and private examples to be less harmful when formatting, terminology, and style are preserved; systematically studying this sensitivity is a useful direction for future work.
    \item \textbf{A Unified Private Evolution Library:} Currently, there is a notable lack of comprehensive frameworks for developing new private evolution (PE) algorithms and corresponding benchmarking pipelines. While~\cite{lin2024differentially} provide an excellent DPSDA library\footnote{\url{https://github.com/microsoft/DPSDA}} for PE, it is built upon Hugging Face's transformers library and lacks direct support for efficient inference libraries like vLLM~\citep{kwon2023efficient}. The computational speedup offered by vLLM over the standard transformers library is substantial; this efficiency gain was the primary motivation for re-implementing our AugPE method using vLLM. Integrating vLLM into the existing DPSDA ecosystem would be a valuable contribution to the community. Additionally, there remains a critical need for robust evaluation pipelines tailored to PE-based algorithms and, more broadly, differentially private (DP) synthetic data generation methods. Although recent efforts have made promising strides in this direction~\citep{wang2025structbench}, the establishment of comprehensive, standardized benchmarks remains an important goal for the community.
    \item \textbf{Privacy Auditing for PE-based Methods:} While PE-based approaches offer rigorous DP guarantees that provide a theoretical upper bound on privacy loss for generated synthetic datasets, empirical privacy auditing remains crucial to establish a practical lower bound. This mirrors the well-established need for auditing in DP model training~\citep{nasr2023tight,steinke2023privacy,cebere2025tighter}. Nevertheless, auditing methods tailored specifically to DP synthetic data generation remain significantly underexplored. Furthermore, the limited existing work on auditing synthetic data privacy~\citep{meeus2025the} is not directly applicable to PE-based methods. Addressing this methodological gap represents an important direction for future research within the community.
\end{itemize}


\section*{Acknowledgments}
We thank Da Yu for the help with the Gemini API settings. We thank Zinan Lin and Chulin Xie for
clarifying questions regarding AugPE. We thank Brendan McMahan and Daniel Ramage for valuable
discussions and feedback on this work.

EC is supported by the Yushan Young Fellow Program (115V1070-1) from the Ministry of Education, Taiwan, NTU Artificial Intelligence Center of Research Excellence (114M7069-01) within Taiwan Centers of Excellence in Artificial Intelligence, and National Science and Technology Council, Taiwan (R.O.C.), under Grant No. NSTC 115-2222-E-002-012.


\bibliography{colm2026_conference}
\bibliographystyle{colm2026_conference}

\appendix

\section{Additional Details of MAPLE}\label{apx:mapledetail}
\subsection{MAPLE procedure}\label{apx:maple_procedure}

\noindent\fbox{%
\begin{minipage}{0.94\linewidth}
\textbf{Inputs:} private text corpus $\mathcal{D}_{\mathrm{priv}}$, non-private donated exemplar set $\mathcal{D}_{\mathrm{don}}$, privacy budget $(\varepsilon,\delta)$, target synthetic dataset size, and PE iteration count $T$.\\[0.4em]
\textbf{Output:} DP synthetic text dataset $\mathcal{D}_{\mathrm{syn}}$.
\begin{enumerate}
    \item Extract structured tabular metadata from each text in $\mathcal{D}_{\mathrm{priv}}$.
    \item Fit AIM on the private metadata and sample DP synthetic metadata using the allocated metadata budget.
    \item For each synthetic metadata record, retrieve matching non-private exemplars from $\mathcal{D}_{\mathrm{don}}$ and construct the \texttt{RANDOM\_API} prompt.
    \item Generate the initial synthetic text dataset conditioned on the DP synthetic metadata and non-private exemplars.
    \item Run $T$ AugPE refinement iterations, using DP nearest-neighbor voting followed by \texttt{VARIATION\_API}.
    \item Return the final synthetic text dataset $\mathcal{D}_{\mathrm{syn}}$.
\end{enumerate}
\end{minipage}}

\subsection{Metadata extraction}\label{apx:step0}
We follow the strategy of~\citet{hu2025actg} to extract metadata in the form of a structured tabular schema. We include the details here for self-completeness. 

\textbf{Schema design.} For the biorxiv dataset, we follow the same schema as in~\citet{hu2025actg} with an additional word count field recording the word count of each text. For the OpenReview dataset, it is inherent with "area" and "recommendation" of the review. We additionally include the following information in the schema: primary contribution type, primary strength, primary weakness, novelty assessment, reviewer tone, clarity and presentation quality, actionability of feedback, word count, and three keywords. For the three keywords, they are chosen from a fixed, disjoint set of keyword pool, respectively. All of them have an option "Other", indicating that none of the keywords in the pool match the review. For the other fields, they all have their respective set of options. Full details can be found at Listing~\ref{lst:meta_biorxiv} and~\ref{lst:meta_openreview} for the biorxiv and OpenReview dataset respectively.

\textbf{Metadata Extraction. }We use \texttt{gemini-2.5-flash-lite} to perform metadata extraction. An example prompt for the biorxiv dataset can be at Listing~\ref{lst:meta_extract_biorxiv}. Note that we assume the data holder can locally host a metadata extractor that does not lead to a privacy breach in our experiment. In practice, if this assumption does not hold, the alternative solution is to leverage privacy-preserving inference methods~\citep{duan2023flocks,hong2024dpopt} for secure API queries or manually annotate the metadata. While this might lead to lower-quality metadata, our ablation study (Figure~\ref{fig:biorxiv_peiter}) shows that even weak metadata still offers significantly better initialization than vanilla AugPE. 

\begin{lstlisting}[caption={Metadata schema for biorxiv dataset}, label={lst:meta_biorxiv}]
attribute_domains = {
  "primary_research_area": domain.CategoricalAttribute([
      "Biochemistry",
      "Bioinformatics",
      "Biophysics",
      "Cancer Biology",
      "Cell Biology",
      "Clinical Trials",
      "Developmental Biology",
      "Ecology",
      "Epidemiology",
      "Evolutionary Biology",
      "Genetics",
      "Genomics",
      "Immunology",
      "Microbiology",
      "Molecular Biology",
      "Neuroscience",
      "Paleontology",
      "Pathology",
      "Pharmacology and Toxicology",
      "Physiology",
      "Plant Biology",
      "Public Health",
      "Scientific Communication and Education",
      "Structural Biology",
      "Synthetic Biology",
      "Systems Biology",
      "Zoology",
      "Other",
  ]),
  "model_organism": domain.CategoricalAttribute([
      "Human",
      "Mouse/Rat",
      "Zebrafish",
      "Drosophila melanogaster",
      "Caenorhabditis elegans",
      "Saccharomyces cerevisiae",
      "Escherichia coli",
      "Arabidopsis thaliana",
      "Plant",
      "Cell Culture",
      "In Silico / Computational",
      "Other Mammal",
      "Other Vertebrate",
      "Other Invertebrate",
      "Other Microbe",
      "Not Applicable / Review",
      "Other",
  ]),
  "experimental_approach": domain.CategoricalAttribute([
      "Wet Lab Experimentation",
      "Computational / In Silico Analysis",
      "Clinical Study",
      "Field Study / Observation",
      "Case Study / Case Review",
      "Review / Meta-analysis",
      "New Method Development",
      "Theoretical Modeling",
      "Other",
  ]),
  "dominant_data_type": domain.CategoricalAttribute([
      "Genomic",
      "Transcriptomic",
      "Proteomic",
      "Metabolomic",
      "Imaging",
      "Structural",
      "Phenotypic / Behavioral",
      "Ecological / Environmental",
      "Clinical / Patient Data",
      "Simulation / Model Output",
      "Multi-omics",
      "Other",
  ]),
  "research_focus_scale": domain.CategoricalAttribute([
      "Molecular",
      "Cellular",
      "Circuit / Network",
      "Tissue / Organ",
      "Organismal",
      "Population",
      "Ecosystem",
      "Multi-scale",
      "Other",
  ]),
  "disease_mention": domain.CategoricalAttribute([
      "Cancer",
      "Neurodegenerative Disease",
      "Infectious Disease",
      "Metabolic Disease",
      "Cardiovascular Disease",
      "Autoimmune / Inflammatory Disease",
      "Psychiatric / Neurological Disorder",
      "Genetic Disorder",
      "No Specific Disease Mentioned",
      "Other",
  ]),
  "sample_size": domain.CategoricalAttribute([
      "Single Subject / Case Study",
      "Small Cohort (<50 subjects)",
      "Medium Cohort (50-1000 subjects)",
      "Large Cohort / Population-scale (>1000 subjects)",
      "Relies on Cell/Animal Replicates",
      "Not Specified / Not Applicable",
  ]),
  "research_goal": domain.CategoricalAttribute([
      "Investigating a mechanism",
      "Characterizing a system/molecule",
      "Developing a method/tool",
      "Identifying novel elements",
      "Testing a hypothesis",
      "Quantifying a parameter",
      "Evaluating/Comparing approaches",
      "Other",
  ]),
  "word_count": domain.CategoricalAttribute(np.arange(50.0,650.0,50)),
}
\end{lstlisting}

\begin{lstlisting}[caption={Metadata schema for OpenReview dataset}, label={lst:meta_openreview}]
attribute_domains = {
  "area": domain.CategoricalAttribute([
      'Deep Learning and representational learning',
      'Applications (eg, speech processing, computer vision, NLP)',
      'Reinforcement Learning (eg, decision and control, planning, hierarchical RL, robotics)',
      'Neuroscience and Cognitive Science (e.g., neural coding, brain-computer interfaces)',
      'Probabilistic Methods (eg, variational inference, causal inference, Gaussian processes)',
      'Social Aspects of Machine Learning (eg, AI safety, fairness, privacy, interpretability, human-AI interaction, ethics)',
      'Unsupervised and Self-supervised learning',
      'Machine Learning for Sciences (eg biology, physics, health sciences, social sciences, climate/sustainability )',
      'General Machine Learning',
      'Theory (eg, control theory, learning theory, algorithmic game theory)',
      'Generative models',
      'Optimization (eg, convex and non-convex optimization)',
  ]),
  "recommendation": domain.CategoricalAttribute([
      '6: marginally above the acceptance threshold',
      '3: reject, not good enough',
      '5: marginally below the acceptance threshold',
      '8: accept, good paper',
      '1: strong reject'
  ]),
  "primary_strength": domain.CategoricalAttribute([
      'Novelty/Originality of Idea', 'Strong Empirical Performance', 'Significant Problem or Application',
      'High-Quality Presentation/Clarity', 'Solid Theoretical Guarantees', 'Good Reproducibility (Code/Data Provided)'
  ]),
  "primary_weakness": domain.CategoricalAttribute([
      'Insufficient/Flawed Experiments', 'Weak or Missing Baselines', 'Lack of Novelty',
      'Unclear/Poor Presentation', 'Flawed Theoretical Claims', 'Limited Scope or Impact', 'Ethical Concerns'
  ]),
  "keywords_group_1": domain.CategoricalAttribute([
      'Large Language Models (LLMs)', 'Diffusion Models', 'Foundation Models', 'Prompting',
      'Text-to-Image Synthesis', 'Transformers', 'Self-Supervised Learning (SSL)',
      'Datasets & Benchmarks', 'Data Augmentation', 'Other'
  ]),
  "keywords_group_2": domain.CategoricalAttribute([
      'Few-Shot & Zero-Shot Learning', 'Meta-Learning', 'Continual Learning', 'Transfer Learning',
      'Federated Learning', 'Graph Neural Networks (GNNs)', 'Knowledge Graphs',
      'Weak Supervision', 'Causal Inference', 'Other'
  ]),
  "keywords_group_3": domain.CategoricalAttribute([
      'Efficiency', 'Stochastic Gradient Descent (SGD)', 'Model Compression', 'Robustness',
      'Out-of-Distribution (OOD) Generalization', 'Fairness & Bias', 'Privacy',
      'Interpretability & Explainability', 'Reinforcement Learning (RL)', 'Decision Making', 'Other'
  ]),
  "word_count": domain.CategoricalAttribute(np.arange(50.0, 1200.0, 50)),
}
\end{lstlisting}

\begin{lstlisting}[caption={Metadata extraction prompt for biorxiv dataset}, label={lst:meta_extract_biorxiv}]
def annotate_features_prompt(
    text: str,
) -> str:
    """Returns a prompt to be used for feature extraction from text.

    This function generates a system prompt for an LLM to annotate a given text
    based on a predefined set of features. The features are derived from a
    Pydantic dataclass. The prompt instructs the LLM to output a JSON object
    containing the annotated features.

    Args:
    dataset_description: A 1-2 sentence description of the dataset.
    dataclass: A Pydantic dataclass defining the features to be extracted.
      Each field in the dataclass represents a feature, and the field's
      type annotation should specify the possible values (e.g., using
      typing.Literal).
    text: The text to be annotated.

    Returns:
    A detailed system prompt suitable for use with frontier language models
    for feature extraction, instructing the model to return a JSON object.
    """
    return textwrap.dedent("""\
      You are an expert biomedical information extraction assistant. Your task is to carefully read a scientific abstract from bioRxiv and extract the specified features according to the schema provided.

Output exactly one JSON object with no extra text or explanations.

**CRITICAL INSTRUCTION 1:** For any field in the JSON schema that lists specific options (e.g., "<Option1|Option2|...>"), you MUST select one of the provided options exactly as it is written. Do not invent, alter, or combine options. Failure to use an exact option from the list will be considered an error.

**CRITICAL INSTRUCTION 2:** Ensure the value chosen for a field is appropriate for that field's specific definition. Do not use an option from one field (e.g., 'Cellular' from `research_focus_scale`) as the value for another field.

Use this schema:
```json
{{
  "primary_research_area": "<Biochemistry|Bioinformatics|Biophysics|Cancer Biology|Cell Biology|Clinical Trials|Developmental Biology|Ecology|Epidemiology|Evolutionary Biology|Genetics|Genomics|Immunology|Microbiology|Molecular Biology|Neuroscience|Paleontology|Pathology|Pharmacology and Toxicology|Physiology|Plant Biology|Public Health|Scientific Communication and Education|Structural Biology|Synthetic Biology|Systems Biology|Zoology|Other>", // Categorizes the abstract into its main biological discipline.
  "model_organism": "<Human|Mouse/Rat|Zebrafish|Drosophila melanogaster|Caenorhabditis elegans|Saccharomyces cerevisiae|Escherichia coli|Arabidopsis thaliana|Plant|Cell Culture|In Silico / Computational|Other Mammal|Other Vertebrate|Other Invertebrate|Other Microbe|Not Applicable / Review|Other>", // Identifies the primary biological model used in the research.
  "experimental_approach": "<Wet Lab Experimentation|Computational / In Silico Analysis|Clinical Study|Field Study / Observation|Case Study / Case Review|Review / Meta-analysis|New Method Development|Theoretical Modeling|Other>", // Describes the main methodology used to conduct the study.
  "dominant_data_type": "<Genomic|Transcriptomic|Proteomic|Metabolomic|Imaging|Structural|Phenotypic / Behavioral|Ecological / Environmental|Clinical / Patient Data|Simulation / Model Output|Multi-omics|Other>", // Specifies the primary type of data generated or analyzed in the paper.
  "research_focus_scale": "<Molecular|Cellular|Circuit / Network|Tissue / Organ|Organismal|Population|Ecosystem|Multi-scale|Other>", // Categorizes the biological level of organization the study focuses on.
  "disease_mention": "<Cancer|Neurodegenerative Disease|Infectious Disease|Metabolic Disease|Cardiovascular Disease|Autoimmune / Inflammatory Disease|Psychiatric / Neurological Disorder|Genetic Disorder|No Specific Disease Mentioned|Other>", // Identifies whether the abstract explicitly names a disease or a major disease category.
  "sample_size": "<Single Subject / Case Study|Small Cohort (<50 subjects)|Medium Cohort (50-1000 subjects)|Large Cohort / Population-scale (>1000 subjects)|Relies on Cell/Animal Replicates|Not Specified / Not Applicable>", // Estimates the scale of the study based on mentions of sample or cohort size.
  "research_goal": "<Investigating a mechanism|Characterizing a system/molecule|Developing a method/tool|Identifying novel elements|Testing a hypothesis|Quantifying a parameter|Evaluating/Comparing approaches|Other>" // Categorizes the study's primary objective based on its framing.
}}
```

**Abstract to analyze:**

{text}

**Your output (JSON only):**
    """).format(
      text=text
    )
\end{lstlisting}

\section{Experiment details}\label{apx:expdetail}

\subsection{Datasets}

We adopt two challenging, real-world datasets for our studies.

\textbf{biorxiv~\citep{hou2025private,hu2025actg} }is a dataset of abstracts on the biorxiv preprint server. We take the processed datasets from~\citet{hu2025actg}. The training set size is $n=28,846$, and the $95\%$ quntile of the context token length is $512$.

\textbf{OpenReview~\citep{xie2024differentially} }is a dataset of paper reviews from ICLR2023, where we take the processed datasets from~\cite{xie2024differentially} directly. The training set size is $n=8396$, and the downstream prediction labels are review area (12 classes) and recommendation (5 classes).

\section{Additional experiments}\label{apx:exp}
\subsection{IMDB initialization results}\label{apx:imdb_init}

We additionally evaluate whether MAPLE provides a stronger initialization on IMDB, a standard benchmark used in prior DP synthetic text work. This study focuses on the initialization stage only, since our goal is to test the same failure mode studied in the main paper: whether unconditional \texttt{RANDOM\_API} generations are poorly aligned with the target data distribution before private evolution begins.

\textbf{Setup.} We use Qwen2.5-7B-Instruct for both MAPLE and AugPE initialization. For MAPLE, we set the total privacy budget to $\varepsilon=3$ and run AIM on IMDB metadata using the allocated metadata budget. The metadata schema contains sentiment, writing style, primary focus, film genre, reviewer expertise, and review length. We then generate a balanced synthetic set of $2{,}000$ reviews, with $1{,}000$ positive and $1{,}000$ negative examples, using DP synthetic metadata and in-context examples retrieved from a pool of $50$ donated examples. For AugPE, we use the standard initialization prompt, ``Write a movie review with a positive/negative sentiment,'' with no metadata or in-context examples; no privacy budget is consumed because this comparison stops before PE voting.

\textbf{Evaluation.} We compare synthetic reviews against a balanced IMDB test reference set of $2{,}000$ reviews using MAUVE. Following the AugPE-style evaluation setup, we use \texttt{stsb-roberta-base-v2} features with scaling factor $5$ and $500$ buckets.

\begin{table}[h]
\centering
\caption{Initialization-only IMDB comparison. Higher MAUVE is better.}
\label{tab:imdb_init}
\begin{tabular}{lcc}
\toprule
Method & Privacy budget used & MAUVE (\%) \\
\midrule
AugPE initialization & none & $1.08$ \\
MAPLE initialization & total $\varepsilon=3$ & $10.43$ \\
\bottomrule
\end{tabular}
\end{table}

MAPLE improves the IMDB initialization MAUVE from $1.08\%$ to $10.43\%$. We also observed that AugPE initialization tends to produce reviews with a concentrated token-length distribution that is far from the real IMDB distribution. MAPLE can incorporate review length as DP metadata, producing a more diverse initial population. These results provide additional evidence that metadata-guided initialization can substantially reduce the initial distribution mismatch before PE refinement.

\end{document}